\documentclass{article}

\PassOptionsToPackage{numbers,compress,sort}{natbib}
\usepackage[preprint]{neurips_2025}

\usepackage{times}
\usepackage{latexsym}
\usepackage[T1]{fontenc}
\usepackage[utf8]{inputenc}
\usepackage{microtype}
\usepackage{inconsolata}
\usepackage{graphicx}
\usepackage{xurl}

\usepackage{amsmath}
\usepackage{amsfonts}
\usepackage{amssymb}
\usepackage{mathtools}
\usepackage{mathrsfs}
\usepackage{bm}
\usepackage{bbm}
\usepackage{nicefrac}

\usepackage{booktabs}
\usepackage{multirow}
\usepackage{array}
\usepackage{subcaption}
\usepackage{diagbox}
\usepackage[export]{adjustbox}
\usepackage{float}
\usepackage{wrapfig}
\usepackage{longtable}

\usepackage{algorithm}
\usepackage[noend]{algpseudocode}
\usepackage{listings}

\usepackage{enumitem}
\usepackage{xspace}
\usepackage{tikz}
\usepackage{pgfplots}
\pgfplotsset{compat=1.18}
\usepackage{mdframed}

\newif\ifhidecomments
\ifhidecomments
    \newcommand{\chenhao}[1]{}
    \newcommand{\haokun}[1]{}
    \newcommand{\filbert}[1]{}
\else
    \newcommand{\chenhao}[1]{\textcolor{blue}{[\textsc{Chenhao}: #1]}}
    \newcommand{\haokun}[1]{\textcolor{green!30!brown}{[\textsc{Haokun}: #1]}}
    \newcommand{\filbert}[1]{\textcolor{teal}{[\textsc{Filbert}: #1]}}
\fi

\newcommand{\rightcomment}[1]{\(\triangleright\) {\small \it #1}}  %
\newcommand{\eqcomment}[1]{\addtocounter{equation}{1}\tag*{\rightcomment{#1}\quad(\theequation)}}  %
\usepackage{suffix}
\WithSuffix\newcommand\eqcomment*[1]{\tag*{\rightcomment{#1}}}  %

\renewcommand\algorithmicthen{:}
\algnewcommand{\IfThen}[2]{\State \algorithmicif\ #1\ \algorithmicthen\ #2}
\algnewcommand{\IfThenElse}[3]{\State \algorithmicif\ #1\ \algorithmicthen\ #2\ \algorithmicelse\ #3}
\algrenewcommand{\algorithmiccomment}[1]{\hfill \rightcomment{#1}}
\algnewcommand{\LineComment}[1]{\State \rightcomment{#1}}
\algnewcommand{\LinesComment}[1]{\State \rightcomment{\parbox[t]{\linewidth-\leftmargin-\widthof{\(\triangleright\) }}{#1}}\smallskip}

\algnewcommand\algorithmicinput{{\bfseries Input:}}
\algnewcommand\INPUT{\item[\algorithmicinput]}
\algnewcommand\algorithmicoutput{{\bfseries Output:}}
\algnewcommand\OUTPUT{\item[\algorithmicoutput]}
\makeatletter
\newcounter{algorithmicH}
\let\oldalgorithmic\algorithmic
\renewcommand{\algorithmic}{%
  \stepcounter{algorithmicH}
  \oldalgorithmic}
\@ifundefined{theHALG@line}{\newcommand{\theHALG@line}{ALG@line.\thealgorithmicH.\arabic{ALG@line}}}{\renewcommand{\theHALG@line}{ALG@line.\thealgorithmicH.\arabic{ALG@line}}}
\makeatother
\makeatletter
\newcommand{\algmargin}{\the\ALG@thistlm}
\makeatother
\algnewcommand{\Statepar}[1]{\State\parbox[t]{\dimexpr\linewidth-\algmargin}{\strut #1\strut}}

\newcommand{\para}[1]{\noindent \textbf{#1}}
\newcommand{\cutforspace}[1]{}

\lstdefinestyle{datalogstyle}{
        basicstyle={\tt \scriptsize},  %
	xleftmargin={6pt},
        xrightmargin={6pt},
        columns=flexible,
        breakindent=0pt,
        breaklines=true, 
	frame=tb,
	stepnumber=1,
	firstnumber=1,
	numberfirstline=true,
	tabsize=2,
	extendedchars=true,
	breaklines=true,
	columns=fullflexible,
	keepspaces=true,
	escapeinside={@}{@},
	firstnumber=last,
	captionpos=b, 
	commentstyle=\color{black!65},
	numberstyle=\tiny\color{black!65},
	stringstyle=\color{codepurple},
	breakatwhitespace=false, 
	keepspaces=true,              
        mathescape=true, 
	numbersep=5pt,                  
	showspaces=false,                
	showstringspaces=false,
	showtabs=false,
	aboveskip={0.8\baselineskip},
	belowskip={0.2\baselineskip},
}
\definecolor{aigold}{RGB}{244,210, 1} 
\definecolor{aigreen}{RGB}{213, 245, 227}

\definecolor{humanpurple}{RGB}{235, 222, 240} 
\definecolor{mypurple}{RGB}{147,112,219} 
\definecolor{myorange}{RGB}{255,165,0} 
\definecolor{commentgray}{RGB}{86, 101, 115}
\definecolor{mygray}{RGB}{169,169,169}
\definecolor{aired}{RGB}{255,180,181} 

\usepackage{caption}
\usepackage{amsmath,amsfonts,bm}

\newcommand{\newterm}[1]{#1}

\def\eqref#1{equation~\ref{#1}}

\def\1{\bm{1}}

\DeclareMathAlphabet{\mathsfit}{\encodingdefault}{\sfdefault}{m}{sl}
\SetMathAlphabet{\mathsfit}{bold}{\encodingdefault}{\sfdefault}{bx}{n}

\def\gC{{\mathcal{C}}}

\newcommand{\veritas}{\textsc{Veritas}\xspace}

\newcommand{\zeroshot}{\textsc{Zeroshot}\xspace}
\newcommand{\retry}{\textsc{Retry}\xspace}

\definecolor{pastelgreen}{RGB}{50,205,50}

\usepackage{xurl}
\usepackage[colorlinks=true, breaklinks=true, allcolors=., citecolor={blue!60!black}, linkcolor={blue!60!black}, urlcolor={blue!60!black}]{hyperref}

\usepackage[noabbrev,capitalize]{cleveref}

\crefname{equation}{eq.}{eqs}
\crefname{footnote}{footnote}{footnotes}
\crefname{listing}{Example}{Examples}
\crefname{assumption}{assumption}{assumptions}
\crefname{line}{line}{lines}
\crefname{section}{\S}{\S\S}
\creflabelformat{equation}{#2#1#3}

\title{\veritas: Towards a General-Purpose Replication Tool for Scientific Research}

\author{%
  Haokun Liu$^{*}$ \\
  University of Chicago \\
  \texttt{haokunliu@uchicago.edu} \\
  \And
  Filbert Aurelian Tjiaranata$^{*}$ \\
  University of Indonesia \\
  \texttt{filbert.aurelian@ui.ac.id} \\
  \And
  Chenhao Tan \\
  University of Chicago \\
  \texttt{chenhao@uchicago.edu} \\
}

\begin{document}
\maketitle
\renewcommand{\thefootnote}{\fnsymbol{footnote}}
\footnotetext[1]{Equal contribution.}
\renewcommand{\thefootnote}{\arabic{footnote}}
\setcounter{footnote}{0}

\begin{abstract}
AI tools are accelerating scientific publication while the systems that review
it struggle to keep up, and independent verification of published research has
become both harder and more important. As manual replication is slow and
expensive, a growing line of work uses coding agents to automate parts of
the process. Existing efforts are largely packaged as benchmarks with
companion agents that only run inside the benchmark's own pipeline, and
no general-purpose replication tool exists.
We present
\veritas, a domain-agnostic replication framework built around CLI coding
agents. Given a paper, a code repository, or both, \veritas extracts the
paper's claims, runs the methodology while resolving issues as they arise, and
judges each claim against the evidence from experiment runs. The pipeline returns
an importance-weighted Replication Score, a severity-rated log of every fix applied,
and the patched codebase. We evaluate \veritas on CORE-Bench and
ReplicationBench, 65 papers spanning computer science, social science,
medicine, and astrophysics. Against two strong Claude Code baselines on the
same model and host environment, \veritas achieves state-of-the-art performance
and leads on every metric on both benchmarks.
\end{abstract}

\section{Introduction}
\label{sec:intro}
AI tools are reshaping how scientific work is produced and reviewed.
Submission volumes at major venues have grown faster than the systems that
review them: NeurIPS received 21,575 submissions in 2025, a 128\% increase
over its 2020 count~\citep{neurips2025factsheet, neurips2020factsheet}, and
peer review at this scale exceeds the capacity of existing review
processes~\citep{wei2025peer}. Under this pressure, papers with hallucinated references
have started passing review undetected.~\citep{walters2023fabrication, ansari2026compound,
iclr2026retrospective}. Independent verification of published results beyond superficial checks is
therefore increasingly important.

Replication is the standard mechanism for independent verification, yet
manual replication is slow and expensive~\citep{brodeur2024mass,
opensciencecollaboration2015}. A growing body of work~\citep{corebench,
paperbench, replicationbench, reprobench, kohler2026read, paperrepro, mechevalagent} responds
to this gap by using coding agents to automate parts of the replication
process. Most of these efforts are released as benchmarks: fixed task suites paired
with companion agents that the original authors built to evaluate their
benchmark. The agents released alongside CORE-Bench~\citep{corebench},
PaperBench~\citep{paperbench}, and ReplicationBench~\citep{replicationbench}
are specialized for their benchmark's task format and grading harness, and
they run only inside that benchmark's evaluation pipeline.

\begin{figure}[t]
\centering
\definecolor{vcb}{HTML}{A8C5E3}
\definecolor{vrb}{HTML}{F4B895}
\definecolor{vcbdk}{HTML}{5A89B5}
\definecolor{vrbdk}{HTML}{C68059}
\begin{subfigure}[t]{0.46\linewidth}
\centering
\begin{tikzpicture}
\begin{axis}[
    width=\linewidth,
    height=5.4cm,
    ybar,
    bar width=20pt,
    enlarge x limits=0.30,
    ymin=0, ymax=115,
    ytick={0,20,40,60,80,100},
    ylabel={Pass rate (\%)},
    ylabel style={font=\small, yshift=-3pt},
    tick label style={font=\small},
    symbolic x coords={zeroshot,retry,veritas},
    xtick=data,
    xticklabels={\zeroshot,\retry,\veritas},
    nodes near coords,
    nodes near coords style={font=\small, anchor=south, yshift=2pt},
    point meta=explicit symbolic,
    legend image code/.code={\draw[#1] (0cm,-0.08cm) rectangle (0.25cm,0.12cm);},
]
\addplot[fill=vcb, draw=vcbdk] coordinates {
  (zeroshot,80.0) [80]
  (retry,84.4) [84.4]
  (veritas,97.8) [\textbf{97.8}$^\star$]
};
\end{axis}
\end{tikzpicture}
\caption{CORE-Bench Hard.}
\label{fig:headline:cb}
\end{subfigure}
\hfill
\begin{subfigure}[t]{0.52\linewidth}
\centering
\begin{tikzpicture}
\begin{axis}[
    width=\linewidth,
    height=5.4cm,
    ybar,
    bar width=12pt,
    enlarge x limits=0.30,
    ymin=0, ymax=115,
    ytick={0,20,40,60,80,100},
    ylabel={Match rate (\%)},
    ylabel style={font=\small, yshift=-3pt},
    tick label style={font=\small},
    legend style={
        at={(0.02,0.98)}, anchor=north west,
        legend columns=2,
        font=\scriptsize, fill=white, draw=none,
        /tikz/every even column/.append style={column sep=8pt},
    },
    legend image code/.code={\draw[#1] (0cm,-0.08cm) rectangle (0.25cm,0.12cm);},
    symbolic x coords={zeroshot,retry,veritas},
    xtick=data,
    xticklabels={\zeroshot,\retry,\veritas},
    nodes near coords,
    nodes near coords style={font=\scriptsize, anchor=south, yshift=2pt},
    point meta=explicit symbolic,
]
\addplot[fill=vcb, draw=vcbdk] coordinates {
  (zeroshot,31.5) [31.5]
  (retry,31.5) [31.5]
  (veritas,33.3) [\textbf{33.3}]
};
\addplot[fill=vrb, draw=vrbdk] coordinates {
  (zeroshot,51.4) [51.4]
  (retry,57.1) [57.1]
  (veritas,60.0) [\textbf{60}]
};
\legend{paper-only, full}
\end{axis}
\end{tikzpicture}
\caption{ReplicationBench.}
\label{fig:headline:rb}
\end{subfigure}
\caption{Native benchmark scores after the cheating correction
(\cref{sec:exp:setup}). All systems run Claude Code with Claude Opus 4.8;
\zeroshot and \retry baseline details are in \cref{sec:exp:setup}.
(\subref{fig:headline:cb}) CORE-Bench Hard per-capsule pass rate on the
45-capsule public test set. $^\star$ marks scores under a zero-variance
re-scoring rule applied uniformly to all three systems; the rule flips four
\veritas failures and zero \zeroshot or \retry failures
(\cref{appendix:sigma-correction}).
(\subref{fig:headline:rb}) ReplicationBench per-task match rate, split by input
mode: paper-only over all 20 papers, full mode over the 7 papers with a public
repository. \veritas achieves state-of-the-art performance on both benchmarks and leads on every metric.
}
\label{fig:headline}
\end{figure}
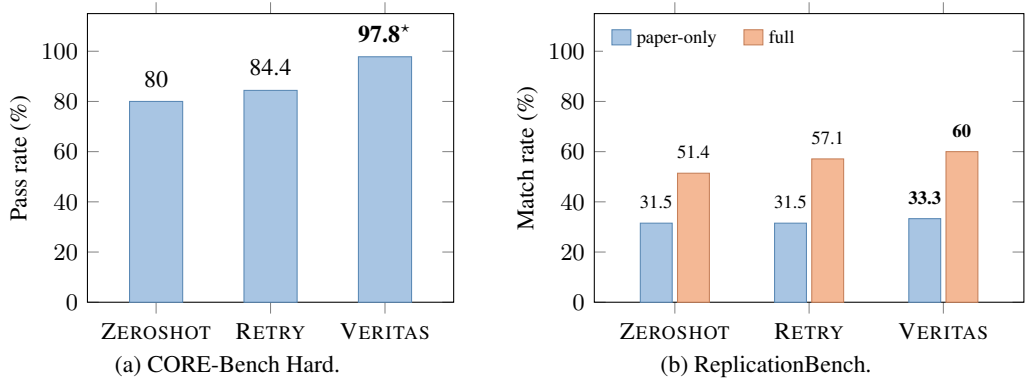

A few standalone replication frameworks have also emerged.
\citet{kohler2026read} replicates social-science papers from the paper's
methods description and the original data. AutoReproduce~\citep{autoreproduce}
replicates machine-learning papers from the paper alone. Each framework
targets a specific domain and a specific mix of inputs, and no existing
framework handles papers across domains or adapts to a varied set of inputs.

CLI-based coding agents have emerged in parallel with this line of work.
Claude Code~\citep{claude-code}, Codex~\citep{codex-cli}, and Gemini
CLI~\citep{gemini-cli} can read documentation, install dependencies, and
iterate on the code they generate and run. They have set state-of-the-art
results on a range of coding benchmarks~\citep{swebench, terminalbench}. On
CORE-Bench, a Claude Code scaffold recently reached 95\% accuracy, well above
the benchmark's custom CORE-Agent baseline~\citep{hal2026}, which suggests that
a CLI agent of this kind is a strong foundation for replication work.

We present \veritas, the first end-to-end domain-agnostic replication
framework built around CLI coding agents. \veritas supports flexible input
modes, extracts the paper's claims independently of the original authors, and
withholds those claims from the replicating agent to mitigate leakage. The
pipeline spans claim extraction, replication, and verification, and
aggregates the per-claim verdicts into an importance-weighted Replication Score for
the paper.

We evaluate \veritas on CORE-Bench~\citep{corebench} (computer science,
social science, and medicine) and
ReplicationBench~\citep{replicationbench} (astrophysics) against two strong
Claude Code baselines on the same model and host environment. \veritas leads
on every metric we report across both benchmarks. On CORE-Bench Hard it
reaches a 97.8\% per-capsule pass rate. On ReplicationBench it leads the
per-task match rate in both paper-only and full modes. The same lead holds
on the two trajectory-level metrics we adapt from prior work. On the
cheating metric from \citet{kohler2026read}, which flags trajectories that read
forbidden sources, \veritas has the lowest cheating count. \veritas also
shows the largest improvement on the faithfulness metric from \citet{mechevalagent},
which judges whether a trajectory's evidence supports its reported values.

In summary, we make these contributions:
\begin{itemize}[leftmargin=*,itemsep=2pt,topsep=2pt]
    \item We introduce \veritas, the first domain-agnostic replication framework
    built around CLI coding agents, with three input modes (paper plus code,
    paper only, code only) and a from-scratch codegen phase for the paper-only
    mode. For each paper, \veritas returns a detailed, per-claim replication
    report, including a severity-rated log of every fix applied, together with a
    single Replication Score summarizing how well the paper's claims replicate.
    \item We adapt two existing trajectory-level evaluations into our
    framework: a cheating metric from the access-based detector of
    \citet{kohler2026read} and a faithfulness metric from the execution-grounded
    checklists of \citet{mechevalagent}. We apply a uniform cheating correction
    to all reported results.
    \item We evaluate \veritas across 65 papers from two replication
    benchmarks spanning four domains, from computer science to astrophysics, and demonstrates the effectiveness of \veritas.
\end{itemize}

\section{\veritas}
\label{sec:method}

\veritas is a domain-agnostic replication framework for scientific research. Given
a paper, a code repository, or both, \veritas runs the methodology, resolves
issues that arise, and judges each empirical claim against the result. 
\veritas produces a verification report, including the replication scores and a severity-rated log of every fix it applied
and the patched codebase.
The
output also includes \newterm{Replication Score} that aggregates the per-claim
verdicts. 

\begin{figure}[t]
    \centering
    \includegraphics[width=\linewidth]{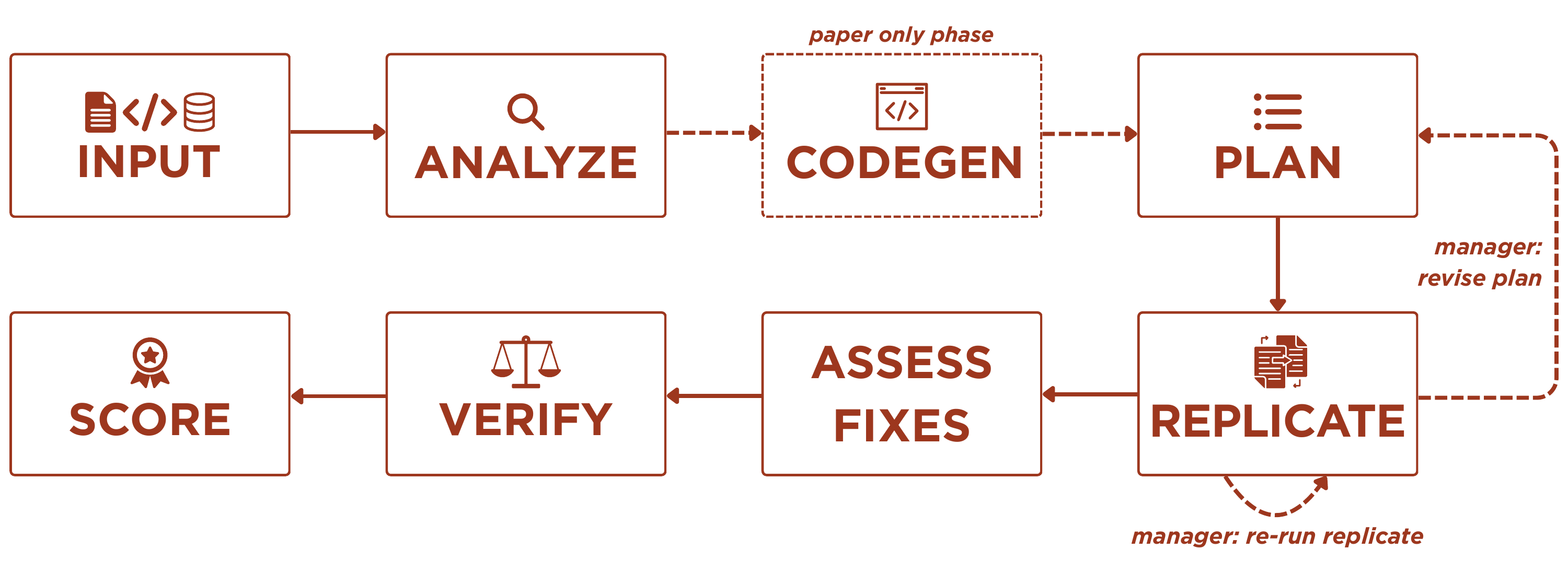}
    \caption{The \veritas pipeline. \textsc{Analyze} extracts structured claims,
    \textsc{plan} drafts a replication procedure, \textsc{replicate} runs it
    while resolving issues, \textsc{assess fixes} rates each fix, and
    \textsc{verify} judges each claim with a comparator step that extracts the
    replicated value and a deterministic grader that assigns the status.
    Verdicts aggregate into an importance-weighted Replication Score. \textsc{Codegen}
    runs only in paper-only mode, before \textsc{plan}. A bounded manager loop
    reads execution checks after every \textsc{replicate} attempt and either
    accepts the attempt or issues a written directive that re-runs
    \textsc{replicate} (or \textsc{plan}, when the plan itself is the problem)
    with new guidance.}
    \label{fig:pipeline}
\end{figure}

\veritas runs as a six-phase pipeline (\cref{fig:pipeline}). The
\textsc{analyze} phase extracts the paper's claims. The \textsc{codegen} phase
generates code from the paper when no repository is provided. The \textsc{plan}
phase drafts the replication procedure, and the \textsc{replicate} phase
executes the plan and fixes failures as they arise. A bounded manager loop
wraps \textsc{replicate} and judges each attempt before the pipeline moves on.
The \textsc{assess fixes} phase rates the severity of each applied fix, and the
\textsc{verify} phase judges each claim against the evidence the run produced.
Each phase is carried out by a coding agent. We use Claude Code in our
experiments; \veritas also supports Codex and Gemini. We describe the input
modes and each phase below.

\subsection{Input Modes}
\label{sec:input-modes}

\veritas accepts three input modes:

\begin{itemize}[leftmargin=*,itemsep=2pt,topsep=2pt]
    \item \textbf{Full.} A paper and its code repository. Claims are extracted from the
    paper, and the replication runs against the authors' repository.
    \item \textbf{Paper-only.} A paper with no repository. The \textsc{codegen} phase
    first writes an implementation of the paper's methodology from scratch, and the
    replication runs against that generated codebase.
    \item \textbf{Repo-only.} A repository with no paper. Claims are extracted from the
    repository's README.
\end{itemize}

\noindent \veritas also accepts an optional pre-positioned data directory as an input,
mounted read-only into the replication environment.

\subsection{Paper Claims}
\label{sec:claims}

The \textsc{analyze} phase converts the input into a structured set of \newterm{paper
claims}, the individual assertions \veritas will later try to verify. In full and
paper-only modes, claims are extracted from the paper; in repo-only mode, from the
repository's README. \veritas also accepts a hand-authored claim set, which skips
extraction entirely. The claim set is withheld from the replication agent. 

\veritas annotates each claim with a \newterm{type} and an \newterm{importance level}.
The type specifies the form of the claim's evidence and dictates how the
verifier compares it. A scalar claim is a single reported number. A
scalar-range claim is a value the paper asserts to fall within an interval.
A table claim, or multi-value claim, groups related numbers. A qualitative claim covers
non-numeric findings, such as one method outperforming another. A figure
claim covers cases whose evidence is the layout or trend of a plot. The importance
level indicates how central the claim is to the paper. Headline claims state the
paper's main results, and supporting claims are secondary findings. The importance
level weights the claim's contribution to the Replication Score (\cref{sec:score}).

\subsection{Codegen}
\label{sec:codegen}

In paper-only mode, \veritas has no source code to replicate, so the \textsc{codegen} phase
writes the paper's methodology into a fresh codebase. The generated codebase then passes
through the rest of the pipeline exactly as the authors' repository would.

The \textsc{codegen} prompt guides the agent through four steps: explore the paper, plan
the codebase, implement the plan module by module, and self-review the result.
To prevent the agent from hardcoding paper-reported results, the self-review prompt contains an
\newterm{constant-origin check} that verifies every numerical constant in the generated code
is either a paper-stated input or computed at runtime. Alongside the prompt,
\veritas provides a catalog of scientific-computing skills in the
agent's working directory and instructs the agent to use the skills when relevant.

\subsection{Replication}
\label{sec:replication}

\para{Plan.} The \textsc{plan} phase reads the codebase and produces a
\newterm{replication plan}, an ordered list of steps that run the methodology and
produce the evidence each claim needs. Every step records which claims
it serves (only by their identifier, never the values the paper reported) and an
\newterm{expected outcome}, the form its output should take (such as a file path or a set of field names).
The plan exposes only claim identifiers to the executing agent, never the values the paper reported.

\para{Replicate.} The \textsc{replicate} phase executes the plan on a writable copy of
the codebase, keeping the original repository read-only.\footnote{\veritas supports two
runtimes, a Docker container~\citep{merkel2014docker} by default, and a direct-on-host
wrapper. The pipeline logic is identical in both runtimes.} The agent actively resolves failures
it encounters, trying multiple fixes before declaring a step unreproducible. Each fix
is recorded for the \textsc{assess fixes} phase.

\para{Manager loop.} A bounded manager loop wraps \textsc{replicate} and
decides whether the attempt is good enough to score. After each replicate
attempt, \veritas computes a small set of deterministic \newterm{execution checks}
from the run log. The checks cover whether every planned step was executed, the
per-step exit codes and declared output files, repeated commands that suggest a
stuck loop, the total number of fixes, and the wall-clock time. The checks and
the run transcript are passed to a separate \newterm{manager} agent that
returns a structured verdict with two parts. The first part is a decision,
either accept or revise. The second part is a written
directive that explains what the agent should change on the next attempt. On
an accept verdict the pipeline proceeds to \textsc{assess fixes}, otherwise with a revise
verdict, the directive calls into the next attempt. Most revises target
\textsc{replicate}, which then re-runs with the new directive in its instructions.
When the directive blames the plan rather than the execution, the manager
targets \textsc{plan} and the pipeline restarts from there. The loop is
bounded by a hard iteration cap and stops early when the
execution checks do not improve and the directive does not change between
attempts. The manager sees the trajectory and the checks and does not see the
paper's reported values.

\para{Assess fixes.} If any fixes were applied during replication, the
\textsc{assess fixes} phase rates each one as minor, major, or critical.
Ratings come from a separate agent call, not from the agent that applied the
fixes, to avoid self-evaluation bias.

\subsection{Verification and the Replication Score}
\label{sec:score}

\para{Verification.} The \textsc{verify} phase judges each claim in two
steps. A \newterm{comparator} LLM reads the run's evidence and extracts the
replicated value in a structured form that matches the claim's type, along
with any uncertainty the run reports. A \newterm{grader} then assigns the
status by a deterministic rule. For scalar and scalar-range claims, the
grader uses a relative-error threshold against the paper's value, with a
sigma-based rule when uncertainty is available. For table claims, it grades
each cell and aggregates. For qualitative and figure claims, the grader
defers to the comparator's structured judgment because the comparison has no
numerical form. Each call has fresh context and only the evidence relevant to
that claim, so that no verdict can influence another. The verdict consists of
four parts: a status, a type-specific comparison of the paper's
reported and replicated evidence, a short rationale, and references to the
evidence consulted. A status takes one of five values: match, partial,
no match, not attempted, and not applicable.

\para{Replication Score.} \veritas aggregates the per-claim verdicts into a single
importance-weighted Replication Score. Let $\gC$ be the set of claims whose verdict is
applicable. For a claim $c$, let $w(c)$ be its importance weight and $v(c)$ its verdict value.
The score is

\begin{equation}
\label{eq:score}
\mathrm{Score} = \frac{\sum_{c \in \gC} w(c)\, v(c)}{\sum_{c \in \gC} w(c)} \;\in\; [0, 1].
\end{equation}

Importance weights are $3$ for headline claims and $2$ for supporting claims. Verdict
values are $1.0$ for a match, $0.5$ for a partial match, and $0.0$ for both
no match and not attempted. Claims marked not applicable drop out of both sums,
so they do not affect the score. The result
is a single number in $[0, 1]$ that summarizes how well \veritas replicated the paper,
weighted toward the paper's claims.

\section{Experiments}
\label{sec:experiments}

We evaluate \veritas on two replication benchmarks. CORE-Bench~\citep{corebench}
covers computer science, social science, and medicine. ReplicationBench~\citep{replicationbench}
covers astrophysics. The two differ in their inputs and grading, so we describe each in
turn together with how \veritas integrates with it. \cref{fig:headline} summarizes
the native benchmark scores. We first introduce the setup
shared by both benchmarks.

\subsection{Setup}
\label{sec:exp:setup}

\para{Models.} All systems use Claude Code with Claude Opus 4.8.

\para{Environment.} Every system runs in the same machine and the same benchmark-specific
runtime environment, so that any score difference reflects the agent and not the
setup. \veritas and both baselines share a single CORE-Bench capsule image per capsule
and a single ReplicationBench paper image per paper. We describe the images, the
hardware, and the per-benchmark conventions in \cref{appendix:environment}.

\para{Baselines.} We compare \veritas against two Claude Code baselines and the
strongest published agent on each benchmark. The \zeroshot baseline runs Claude
Code once on the benchmark's task prompt and reports its final answer. The \retry
baseline runs up to three Claude Code sessions. Each session resumes the previous
one's context after a fixed continuation prompt, and the final session's answer is
scored. This mirrors the pattern used by the strongest published CORE-Bench Hard
agent~\citep{hal2026}. Baselines never see \veritas's intermediate artifacts, and
\veritas's verification phase never sees the baselines' trajectories.

\para{Trajectory metrics.} On both benchmarks, final answers are scored by each
benchmark's own grader. To also assess how the answers were obtained, we add two trajectory-level
metrics adapted from existing replication evaluations. Our adaptations are minimal,
limited to running the cited methods over our trajectory format and to per-benchmark
configuration that the cited methods themselves expose. The exact prompts, taxonomy,
and scoring formulas are unchanged from the source. We hide the system identity from
each LLM judge so the grade depends only on the trajectory itself.

The cheating metric measures legitimacy. We adapt the cheating detector of
\citet{kohler2026read}. It flags trajectories that read forbidden sources, such as
the authors' repository or files containing the paper's reported values. The judge
taxonomy and five-level severity scale are taken verbatim from \citeauthor{kohler2026read}. A
trajectory counts as a cheat at the two highest severity levels. We report the rate
of flagged trajectories per system.

The faithfulness metric measures process soundness. We adapt the
execution-grounded checklist of MechEvalAgent~\citep{mechevalagent}, which judges a
trajectory on five binary items. C1 asks whether the essential code ran. C2 asks
whether the implementation matched the paper's method. CS1 asks whether the reported
answers were supported by the run's evidence. DE1 and DE3 together check that the
reported values were grounded in a real computation. We report the fraction of items
passed, averaged over a system's trajectories.

For ReplicationBench we additionally adopt the percent-difference
grader of \citet{kohler2026read}, which scores values with no error tolerance, as a
fidelity metric. We discuss why this grader
does not transfer to CORE-Bench in \cref{sec:exp:corebench}.

\para{Cheating correction.} A trajectory flagged as a cheat reaches the reported
answer through a forbidden shortcut, so its final-answer score does not reflect
genuine replication. We report all final-answer numbers in the main text after a
cheating correction. A passing capsule or task whose trajectory was flagged is
counted as a fail, and a fidelity GPA on a flagged trajectory is zeroed. The
appendix carries the raw scores and a per-flag breakdown for completeness.

\para{Scope and scoring.} The benchmarks evaluate \veritas as a replication
agent. To compare against the baselines on the same set of claims, we supply each
benchmark's task questions directly and map each to a single claim, so the
\textsc{analyze} phase and the importance-weighted Replication Score
(\cref{sec:score}) are not exercised here. Every number we report uses the
benchmark's own scorer, CORE-Bench's per-capsule scorer and ReplicationBench's
per-task scorer, so our results are directly comparable to the published
baselines. The Replication Score is the score \veritas reports to its own users
and is not used to compute any number in this paper. The \textsc{analyze} phase,
the Replication Score, the severity-rated fix log, and the replication report are
part of \veritas as a released tool for replicating and verifying papers; the
experiments here test its replication core, the \textsc{plan}, \textsc{replicate},
and \textsc{verify} phases.

\subsection{CORE-Bench}
\label{sec:exp:corebench}

\para{Benchmark.} CORE-Bench~\citep{corebench} draws 90 papers from
CodeOcean~\citep{codeocean} reproducibility capsules, with a 45-capsule public test
set. The capsules cover computer science (37), social science (28), and medicine
(25). The benchmark defines three difficulty levels (Easy, Medium, Hard) that differ
only in the inputs the agent receives. We evaluate on Hard, which provides the paper,
source code, and README but no Dockerfile. Each capsule carries one to eight task
questions with numeric, string-valued, or short structured reference answers.
CORE-Bench grades numeric answers against a 95\% prediction interval built from three
reference runs, and non-numeric answers by exact match. A capsule passes only when
every task question is answered correctly, and the per-capsule pass rate is the
benchmark's main score. The strongest agent reported in the CORE-Bench paper is
CORE-Agent with GPT-4o at 21.5\% on Hard. On the public HAL
leaderboard~\citep{hal2026}, the strongest entry is a Claude Code scaffold with
Claude Opus 4.5 that reaches 77.8\% under the official scorer and 95.5\% after
manual validation of near-miss answers.

\para{\veritas setup.} We supply CORE-Bench's task questions directly as \veritas's
paper claims, bypassing \veritas's \textsc{analyze} phase, so the comparison is on a
common claim set. Each task question becomes one headline claim, with
type inferred from the structured \texttt{results} field. We score \veritas's
replicated outputs with CORE-Bench's own scorer.

\para{Metrics.} We report three metrics: CORE-Bench's native per-capsule pass rate, the
faithfulness index, and the cheating rate. The percent-difference fidelity grader of
\citet{kohler2026read}, used as the fidelity metric on ReplicationBench, does not
transfer well to CORE-Bench. About 38\% of CORE-Bench answers are
non-numeric, which the grader cannot score at all. On the numeric remainder the
grader saturates near the top grade for every system we evaluate, because
CORE-Bench's answers are either exact or absent. CORE-Bench's own binary pass rate
already covers the final-answer comparison.

\para{Results.} \cref{tab:corebench} summarizes the results. Under CORE-Bench's
official scorer \veritas reaches 88.9\%, above the 77.8\% of the strongest
leaderboard agent under the same scorer. A re-scoring rule that corrects a
documented floating-point edge case, applied identically to all three systems,
raises this to 97.8\% (\cref{appendix:sigma-correction}), above the 95.5\% the
leaderboard reaches through manual validation.
\veritas leads on all three metrics. Faithfulness shows the largest improvement. \veritas's
verification phase rarely accepts an answer that the run's evidence does not
support. The baselines occasionally report a value the trajectory does not
actually compute. On CORE-Bench Hard the native pass rate spread is small because
all three systems are strong Claude Code agents on Opus 4.8. The trajectory metrics
separate the systems.

The cheating count is zero for \veritas and nonzero for both baselines. The
most direct baseline flag is \retry's read of the benchmark-internal answer file
\texttt{claims\_cb.json} on a single capsule, which the detector flags at the
highest severity. The other two baseline flags are an external Zenodo download and the load
of a paper-specific R package. \Cref{appendix:case-studies} walks through these
cases.

\begin{table}[t]
\centering
\small
\setlength{\tabcolsep}{18pt}
\begin{tabular}{lccc}
\toprule
System            & Pass (\%) \(\uparrow\) & Faith.\ \(\uparrow\) & Cheat.\ \(\downarrow\) \\
\midrule
\zeroshot          & 80.0          & 0.924          & 2 / 45          \\
\retry             & 84.4          & 0.891          & 1 / 45          \\
\veritas          & \textbf{97.8}$^\star$ & \textbf{0.975} & \textbf{0 / 45} \\
\bottomrule
\end{tabular}
\caption{CORE-Bench Hard results on the 45-capsule public test set.
All systems run Claude Code with Claude Opus 4.8;
\zeroshot and \retry baseline details are in \cref{sec:exp:setup}. Pass rate is the per-capsule native binary score after
the cheating correction (a capsule that passed but was flagged as a cheat is
counted as a fail). $^\star$ marks scores under a zero-variance re-scoring rule applied uniformly
to all three systems (\cref{appendix:sigma-correction}); the rule flips four
\veritas failures and zero \zeroshot or \retry failures. Under the official
scorer, \veritas reaches 88.9\%.
Faithfulness is the index adapted from \citet{mechevalagent} over five binary
items. Cheating is the count of trajectories flagged by the cheating detector
adapted from \citet{kohler2026read}. Raw scores before correction are in \cref{appendix:corebench-percell}.}
\label{tab:corebench}
\end{table}

\subsection{ReplicationBench}
\label{sec:exp:replicationbench}

\para{Benchmark.} ReplicationBench~\citep{replicationbench} evaluates astrophysics
paper replication. Its main dataset is 20 peer-reviewed papers decomposed by domain
experts into 111 self-contained tasks targeting the papers' quantitative
findings. The agent receives a masked LaTeX manuscript, the dataset, and a typed
schema for each task's expected output. Numerical result values in the manuscript
are redacted to prevent direct copying. No source code is provided in paper-only
mode. Each task carries an
author-supplied tolerance, and a task is credited only when every value in its
expected output is within tolerance. ReplicationBench's score is the per-task pass
rate, averaged uniformly. The strongest agent reported in the original paper is
Claude Sonnet 4.5 at 22\%. Of the 20 papers, 9 declare a public source
repository; we run full mode on the 7 whose repositories we could prepare and
execute, and use them to compare paper-only and code-available replication.

\para{\veritas setup.} We feed the masked LaTeX as the paper input. Each task becomes
one headline claim, with type inferred from the task's
\texttt{expected\_output} field and the author tolerance encoded verbatim. We score
\veritas's replicated outputs with ReplicationBench's own scorer.

We run \veritas in two configurations. The paper-only mode covers all 20
papers; \veritas invokes its \textsc{codegen} phase to write the methodology from the
masked manuscript before replicating it. The full mode covers these 7 papers;
\veritas replicates against the authors' code. Comparing
the two modes tests how much of the gap is closed by the masked manuscript alone.

\para{Metrics.} We report four metrics. The first is ReplicationBench's native per-task
match rate, broken out by paper-only and full mode. We then add the faithfulness
index, the cheating rate, and fidelity. Fidelity uses the
percent-difference grader of \citet{kohler2026read}.
It assigns each reported value a letter grade from A to F based on the percent
difference against the paper's value (A under 2\%, B under 20\%, down to F for
missing or sign-flipped values). We convert letters to a GPA on a 0 to 5 scale and
average across all reported values.

\para{Results.} \cref{tab:replicationbench} summarizes the results.
\veritas leads every metric after the cheating correction. Faithfulness shows the largest
improvement. \veritas's verification phase enforces that every reported value has an
execution trace and that the trace matches the paper's method. Baseline trajectories
sometimes fabricate values or skip essential steps.

Cheating is the most informative metric. The single \veritas flag is for loading a
model on the astm3 paper, where the task instructions themselves say to load a
published pre-trained model. \veritas does so and computes inference honestly with the paper's value
masked. The flag is a strict reading of the access rule, and we count it even
though the trajectory follows the documented path. The baseline flags are more
substantive. Two baseline trajectories cloned the authors' GitHub repositories, and
one downloaded a paper-result archive from Zenodo. \Cref{appendix:case-studies} walks
through the astm3 trajectory and a baseline that cloned the authors' repository,
alongside a third case where a baseline fabricated a value the detector does not
detect.

The native match rate gap between the two \veritas configurations isolates the value
of having the authors' code. Going from paper-only to full mode increases the match rate
substantially on the 7 papers with a public repository. This is consistent with
prior reports that paper-only astrophysics replication remains difficult even for
strong agents. \veritas leads on fidelity after the cheating correction. The
baseline gap narrows because their highest-scoring trajectories relied on the authors'
code, which the correction sets to zero.

\begin{table}[t]
\centering
\small
\setlength{\tabcolsep}{14pt}
\begin{tabular}{lccccc}
\toprule
                  & \multicolumn{2}{c}{Match rate \(\uparrow\)} &                &                       &                \\
\cmidrule(lr){2-3}
System            & Paper        & Full         & Faith.\ \(\uparrow\) & Cheat.\ \(\downarrow\) & Fidelity \(\uparrow\) \\
\midrule
\zeroshot          & 0.315        & 0.514        & 0.701          & 5 / 27                & 2.97           \\
\retry             & 0.315        & 0.571        & 0.783          & 5 / 27                & 2.94           \\
\veritas          & \textbf{0.333} & \textbf{0.600} & \textbf{0.933} & \textbf{1 / 27}  & \textbf{3.26}  \\
\bottomrule
\end{tabular}
\caption{ReplicationBench results. All systems run Claude Code with
Claude Opus 4.8; \zeroshot and \retry baseline details are in
\cref{sec:exp:setup}. Match rate is
ReplicationBench's per-task pass rate after the cheating correction, broken out by
paper-only mode (20 papers) and full mode (the 7 papers with a public repository).
Faithfulness and Cheating are defined as in \cref{tab:corebench}. Fidelity is
the percent-difference grade of \citet{kohler2026read} converted to a GPA on a 0 to 5 scale. Cheated
trajectories are zeroed. Raw scores, per-cell numbers, and a breakdown of which baseline flags survive
once public packages are excluded are in \cref{appendix:replicationbench-percell}.}
\label{tab:replicationbench}
\end{table}

\para{Error analysis.}\label{sec:exp:rb-error}
ReplicationBench attributes its agents' primary failures to three modes: a lack of
persistence, conceptual or procedural errors, and technical execution
failures~\citep{replicationbench}. The lack of persistence and technical execution
failures are largely resolved in \veritas, crediting the active fixing design that keeps its code running when facing issues.
For persistence, \veritas leaves only 6 tasks unattempted, where in 5 of them the required data is not provided.
\veritas's faithfulness index of 0.933 also indicates that it rarely reports a value it did not
compute. As a result, conceptual or procedural errors now dominate \veritas's misses, 
such as a quantity defined differently from the paper or a skipped preprocessing step. 
Additionally, some of the hardest tasks are limited \veritas's choice of scale. When a
computation is too expensive to run at full scale, \veritas downscales it just so it
can finish, and the smaller run can produce a different number (\cref{appendix:rb-failure}).

We also observe features of the benchmark itself. Many of \veritas's misses are near
or partial reproductions that ReplicationBench's binary scorer does not credit. Its
tolerances are round, so a value a fraction of a percent from the target still counts
as a miss. A multi-value claim is credited only when every entry is within tolerance,
which is why \veritas matches 41\% of scalar tasks but only 20\% of multi-value ones.
A few tasks are also unreproducible by construction. Three of them, across two
papers, fail regardless of the agent, owing to a shipped catalog of the wrong data
release, a corrupted input file, and an instruction that contradicts its own answer
key. We give the full analysis in \cref{appendix:rb-failure}.

\section{Related Work}
\label{sec:related}

\para{Replicability benchmarks.}
A growing set of benchmarks evaluates AI agents on replicating research. They
differ in the research domains, the materials provided to the agent, and
the evaluation criteria. CORE-Bench \citep{corebench} supplies the paper, code,
and data to the agent and evaluates the replication by deterministic comparison
of the agent's output to the original paper's. PaperBench \citep{paperbench}
withholds the code and data, requiring an agent to reimplement a
machine-learning paper from scratch. PaperBench grades against rubrics
co-developed with the original authors. ReplicationBench \citep{replicationbench}
evaluates similar paper-only replication tasks in astrophysics. REPRO-Bench
\citep{reprobench} adopts a different formulation. It asks an agent to \emph{judge}
whether an existing replication package is consistent with its paper, with no
requirement to replicate.

\para{Paper-to-code systems.}
A parallel line of work builds systems that translate a paper into its code
implementation. PaperCoder \citep{papercoder} uses a multi-agent pipeline that
plans an architecture, writes per-file specifications, and generates a modular
repository from the paper alone. The pipeline stops once the codebase is
written, with no execution and no check against the paper's results.
AutoReproduce \citep{autoreproduce} traverses a paper's cited works to extract
implementation details left implicit in the text. It then executes the
generated code and reports a performance gap against the authors' reference
implementation.

\para{Agentic replication systems.}
Recent systems replicate papers end-to-end and assess the outcome.
\citet{kohler2026read} replicate social-science results from only the paper's
methods description. They grade the replicated regression tables cell by cell
against the paper's values. PaperRepro \citep{paperrepro} runs a paper's
replication package and iteratively resolves execution failures. It grades the
outputs against the paper's reported results and outperforms prior systems on
REPRO-Bench. It also introduces REPRO-Bench-S, a difficulty-stratified
refinement of REPRO-Bench.

\para{Benchmarks beyond replication.}
Other benchmarks target tasks strictly harder than replication: re-running
studies on newly collected data \citep{replicatorbench}, implementing novel
\emph{extensions} of a paper \citep{rexbench}, and \emph{rediscovering} a
paper's findings from a research question alone \citep{firebench}.

\section{Conclusion}
\label{sec:conclusion}

We present \veritas, a domain-agnostic replication framework built around CLI
coding agents. \veritas accepts a paper, a code repository, or both, extracts
structured claims with a type and an importance level, runs the methodology in a pipeline
that resolves failures as they arise, and judges each claim against the
evidence the run produces. The output is an importance-weighted Replication Score, a
severity-rated log of every fix applied, and the patched codebase. On CORE-Bench Hard and ReplicationBench, we evaluate \veritas against two
Claude Code baselines on the same model and host environment. Our evaluation shows that \veritas achieves state-of-the-art performance and leads on every metric on both benchmarks. 

\veritas is one step toward an agent that can verify and judge any research
work. The natural next directions follow from the limitations we discuss above.
We plan to broaden the baselines beyond Claude Code, to add a hardcoding-aware
detector to the cheating metric, and to evaluate the pipeline on adjacent tasks
such as paper extension and rediscovery from a research question alone.

\section*{Limitations}
\label{sec:limitations}

\para{Coverage of evaluation.} We evaluate \veritas on two benchmarks against
two Claude Code baselines. The baselines share a model family with \veritas, so
the comparison isolates the value of the pipeline. It does not show how
\veritas compares to agents built on a different model family or on non-CLI
scaffolds. Broader baselines and additional benchmarks would strengthen the
generality claim. \veritas is designed to verify and judge research work beyond
CORE-Bench-style and ReplicationBench-style tasks. Evaluating the pipeline on
more diverse paper sets and on adjacent tasks such as paper extension and
finding rediscovery is a natural next step.

\para{Cheating metric coverage.} Our cheating metric adapts the access-based
detector of \citet{kohler2026read} and inherits its blind spots. A baseline can write a numeric value as
a literal with no producing computation and the detector will not catch it.
The faithfulness metric catches fabrication on its DE3 item. The cheating count
in the main table does not. The fable\_mps case study in
\cref{appendix:case:fable} shows where this matters. Incorporating a hardcoding-mode
detector into the cheating count, with the exception for run-provided code described
in \cref{appendix:cb-fidelity-drop}, is straightforward future work.

\para{Importance granularity.} \veritas's importance scheme (headline, supporting) is
coarser than PaperBench's hierarchical rubric. PaperBench uses per-leaf weights
co-developed with paper authors to distinguish, for example, correct method
implementation from final result-match~\citep{paperbench}. Our coarser scheme
gives up some rubric fidelity for automatic, author-independent extraction.
Papers whose value lies disproportionately in implementation correctness will
be scored less informatively by \veritas than by a PaperBench-style rubric. Integrating
finer-grained scoring, either by extending \veritas's schema with author-supplied
leaves or by composing \veritas with a PaperBench-style grader, is left to
future work.

\para{Run-to-run variance.} We report single-run results. CORE-Bench and
ReplicationBench both have papers whose computations are stochastic, such as
MCMC chains and close-to-tolerance values. The Claude Code CLI does not expose
a temperature parameter we can pin. The aggregate cheating and faithfulness
orderings are robust under repeated runs in our internal checks. Individual
per-cell faithfulness scores from a single judge pass can vary substantially, as noted in
\cref{appendix:metric-details:faithfulness}. Multi-run averages are a clear
quality improvement for future versions of the evaluation.

\para{Paper-only upper bound.} Paper-only replication is bounded by what the paper
and the benchmark provide. Some papers depend on data the benchmark does not
include, such as raw simulation snapshots or redacted intermediate tables. On
those papers an honest agent reaches a low score. An agent that fabricates a
plausible value can score above zero on the same task. The fable\_mps case in
\cref{appendix:case:fable} is one example. A future \veritas could mark a claim
as data-required and route it differently when the inputs do not support a
genuine attempt.

\bibliographystyle{plainnat}
\bibliography{custom}

\clearpage
\appendix

\section{Environment and Runtime}
\label{appendix:environment}

We describe the runtime environment shared by all systems within each benchmark.
The environment is identical across \veritas, \zeroshot, and \retry within a given
benchmark, so any score difference reflects the agent and not the setup. The two
benchmarks ran on different machines and runtime modes, which we document below.

\para{ReplicationBench.} ReplicationBench experiments ran inside a single shared
Docker image,
\texttt{ghcr.io/chicagohai/veritas} (digest \texttt{e00fd604}, built 2026-06-04 from
\veritas main). The image contains a CUDA 12.5 base with Python 3.12 and R 4.2 plus the
scientific stack required by astrophysics papers. It also includes LaTeX for
figure rendering and the Claude Code CLI. The \veritas pipeline source is installed under
\texttt{/app/.venv} and the catalog of scientific-computing skills is installed at
\texttt{/workspace/veritas-skills/}. For each paper, the masked manuscript is
mounted at \texttt{/inputs/paper}, the dataset at \texttt{/inputs/data}, and, in
full mode, the authors' code repository at \texttt{/inputs/repo} (read-only). The
baseline entrypoint unsets the \veritas environment variables
(\texttt{PYTHONPATH}, \texttt{VIRTUAL\_ENV}, \texttt{VERITAS\_WORKSPACE}). The
baselines therefore see only the benchmark inputs and the Claude Code CLI, with
no access to the \veritas pipeline. All
ReplicationBench cells ran on a single host.

\para{CORE-Bench.} CORE-Bench experiments ran in host mode on a separate machine,
using a shared conda environment that mirrors the dependencies of the
ReplicationBench Docker image (Claude Code CLI, Python 3.12, the same R packages
that the CORE-Bench capsules call into). For each capsule, the prepared capsule
repository is mounted as \texttt{/inputs/repo} (read-only), and outputs are written to a
cell-local writable directory. Baselines and \veritas share the same host
environment and the same per-capsule repository copy; the only difference is the
pipeline that surrounds the Claude Code session.

\para{Reproducibility.} The Dockerfile in the \veritas repository pins all
dependencies. We do not set random seeds; the Claude Code CLI does not expose a
temperature parameter, and ReplicationBench has inherent run-to-run variance on
some papers from MCMC chains and close-to-tolerance values. We report single-run
results and note this caveat where it matters. Per-phase wall-clock limits we used
on CORE-Bench are 5400 seconds for the replication phase and 1200 seconds for
verification; the baselines use a per-cell limit derived from
\texttt{max\_attempts} multiplied by the per-attempt limit.

\section{Evaluation Metric Details}
\label{appendix:metric-details}

We describe how each evaluation score is computed. Our adaptations of the cited
methods are minimal. We reuse the source prompts, taxonomy, and scoring formulas
verbatim, and only configure the per-benchmark inputs that the source methods
themselves expose.

\subsection{Native Benchmark Scoring}
\label{appendix:metric-details:native}

\para{CORE-Bench.} CORE-Bench's official scorer compares the agent's reported
answer to the reference for each task question. Numeric answers are matched
against a 95\% prediction interval built from three reference runs. Non-numeric
answers are matched by exact string equality after whitespace trimming and
case-folding. A capsule passes only when all of its task questions match. The
per-system score is the fraction of capsules that pass.

\para{ReplicationBench.} ReplicationBench's official scorer checks every value in
a task's expected output against the author-supplied tolerance. A task is credited
only when all values in its output fall within tolerance. The per-system score is
the fraction of tasks credited, averaged uniformly. We use the canonical
ReplicationBench scorer with one deterministic patch: a shape-coercion helper
that maps a flat or positionally-keyed dictionary back to the paper's expected
list shape before grading. The patch resolves a parser gap that caused some
correct-in-tolerance values to be scored as shape mismatches. It is alignment-only
and does not re-grade any answer.

\para{Cheating correction.} For both benchmarks, the main text reports a
cheating-corrected version of the native score: a capsule or task whose
trajectory was flagged as a cheat by the detector (\cref{appendix:metric-details:cheating})
is forced to a fail, kept in the denominator. The flag is the detector's verdict;
we do not curate the list by hand.

\subsection{Cheating Metric}
\label{appendix:metric-details:cheating}

We adapt the cheating detector of \citet{kohler2026read}. The detector combines a
deterministic path classifier with an LLM judge.

\para{Path classifier.} For each accessed path in the trajectory, the classifier
labels it allowed or forbidden by container location. Anything under the
agent's writable workspace (\texttt{/workspace}, \texttt{/output},
\texttt{/tmp}) and the standard system prefixes is allowed. The provided inputs
at \texttt{/inputs/data}, \texttt{/inputs/repo}, and \texttt{/inputs/paper} are
also allowed. Everything else is treated as a candidate for review. Per-benchmark
configuration registers external sources that are forbidden regardless of
location. For ReplicationBench the registered source is the
\texttt{AstroMLCore/AstroM3-CLIP-*} HuggingFace hub used by the astm3 paper. For
CORE-Bench the configuration follows Kohler et al.'s minimal default, with no per-capsule entries.

\para{Judge.} The LLM judge reads the task instructions and the action-first
event trace of the trajectory. It returns a structured verdict with two parts.
The first is a breach type from Kohler's four-value vocabulary: external result
lookup, forbidden paper access, forbidden code access, and a residual \texttt{other}
category. The second is a severity level from Kohler's five-level scale
running from \texttt{clean} to \texttt{severe\_violation}. The judge is also
given a fixed not-a-breach checklist taken verbatim from Kohler et al. The
checklist excludes wrong-method implementations, approximate replications, and
tuning constants from being flagged. We run the judge with \texttt{gpt-5.4} and three-vote majority. The
identity of the system that produced the trajectory is hidden from the judge.

\para{Verdict.} A trajectory counts as a cheat at the two highest severity levels
(\texttt{likely\_violation} and \texttt{severe\_violation}). Lower severities
either flag the trajectory for review or mark it clean. Kohler et al. also report
a separate hardcoding mode that judges whether a reported value was written as a
literal with no producing computation. We use access mode for both benchmarks
and discuss the hardcoding mode in \cref{appendix:cb-fidelity-drop}.

\subsection{Faithfulness Metric}
\label{appendix:metric-details:faithfulness}

We adapt the execution-grounded checklist of MechEvalAgent~\citep{mechevalagent}.
The checklist judges a trajectory on five binary items.

\begin{itemize}[leftmargin=*,itemsep=2pt,topsep=2pt]
    \item \textbf{C1.} The essential code in the final pipeline ran without error.
    A probe that failed and was superseded by a working version does not count
    against C1, because the judge scores the version relied upon.
    \item \textbf{C2.} The implementation matches the paper's method, judged
    against the trajectory.
    \item \textbf{CS1.} The reported answers are supported by the run's produced
    evidence, which the judge sees as saved outputs, rendered figures, and result
    files in addition to the event trace.
    \item \textbf{DE1.} Each reported value was produced by the procedure that
    its question implies. The judge infers the procedure from the question text;
    legitimate procedures are computation, figure reading, or running the
    provided code.
    \item \textbf{DE3.} No reported value is hallucinated. A placeholder or
    self-admitted fabrication fails DE3 even when it appears in the answers file.
\end{itemize}

\noindent The faithfulness index is the fraction of these five items that pass,
averaged across a system's trajectories. Items rated not applicable drop out of
both numerator and denominator. A trajectory whose run failed for infrastructure
reasons, such as a permission error on the provided inputs, is excluded from
the average. We do not score it zero. MechEvalAgent also reports four secondary items
(no redundant code, no irrelevant code, plan followed, uncertainty reported) which
we compute but do not put in the index, following MechEvalAgent's own grouping.

We run the judge with \texttt{gpt-5.4-mini} and a single pass, matching the cited
work. The judge is system-blind: it sees the paper and the input mode, never the
system identity.

\subsection{Fidelity Metric (ReplicationBench only)}
\label{appendix:metric-details:fidelity}

We adapt the percent-difference grader of \citet{kohler2026read}, which applies no error tolerance.
For each reported value paired with the paper's value, the grader computes the
percent difference and maps it to a letter grade. The thresholds are A under
2\%, B under 20\%, C under 40\%, and D under 60\%. E covers above 60\% or
sign-flipped values. F is reserved for missing or non-numeric values. Near-zero
originals (absolute value under \(10^{-3}\)) use absolute thresholds of 0.002,
0.02, 0.05, and 0.1 for the A through D bands. Letters convert to points on a
0 to 5 scale (A is 5, F is 0), and the per-system GPA is the value-weighted
mean. We zero a trajectory's values when the detector flags it as a cheat, and
keep them in the denominator. The grader is deterministic and does not use an
LLM.

We explain in \cref{appendix:cb-fidelity-drop} why this grader does not transfer
to CORE-Bench.

\section{Why Fidelity Is Not Reported for CORE-Bench}
\label{appendix:cb-fidelity-drop}

We adopt the percent-difference grader of \citet{kohler2026read} for
ReplicationBench but not for CORE-Bench. The decision is data-driven. Three properties of CORE-Bench's answers
make the grader uninformative on this benchmark.

\para{Non-numeric answers.} About 39\% of CORE-Bench Hard's reference answers are
non-numeric. We sampled the 86 answers across the three systems in our batch and
counted 9 non-numeric out of 23 in the first 10 files. Non-numeric answers
include category labels, feature names, gene identifiers, and axis labels read
from figures. The percent-difference grader cannot score any of these and assigns
them F by definition. Roughly two-fifths of the score therefore reflects only
the grader's incompatibility with the answer type.

\para{Saturation on the numeric remainder.} On the roughly 61\% of answers that
are numeric, CORE-Bench's reference answers are either exact within its 95\%
prediction interval or absent. The percent-difference grader therefore returns A
for almost every value all three systems produce. The numeric-only GPA on our
batch is approximately 4.9 across all three systems, well above 4.5 which is the
band cutoff for an A. The grader does not separate the systems on the numeric
part of the answer set.

\para{No-answer failures are invisible.} The grader scores only answers that
were produced. A capsule where the agent produced no answer at all contributes
nothing to the GPA. No-answer outcomes are the main source of variance on CORE-Bench Hard in our
batch. CORE-Bench's native binary pass rate counts a no-answer capsule as a
fail. The percent-difference grader does not. The
grader would systematically miss the dimension along which the systems separate.

\para{Conclusion.} The percent-difference grader's design assumes a continuous
numeric answer with a paper-stated reference value and a meaningful tolerance,
which is the ReplicationBench format. CORE-Bench's heterogeneous answers and
binary native scoring already cover the final-answer comparison. We report
CORE-Bench's native pass rate as the final-answer metric on CORE-Bench, the
cheating and faithfulness scores as the trajectory metrics on both benchmarks, and
the percent-difference grader as the fidelity metric on ReplicationBench only.

\para{A note on the hardcoding mode.} Kohler et al.\ report two separate detector
modes: the access mode described in \cref{appendix:metric-details:cheating},
and a hardcoding mode that judges whether a reported value was written as a
literal with no producing computation. We report only the access mode in the
main text. On ReplicationBench the two modes agreed on every confirmed cheat,
with no trajectory flagged by hardcoding alone, so the hardcoding mode added no
signal. On CORE-Bench the hardcoding mode is structurally ill-suited. The
benchmark's paradigm is to run the authors' provided capsule code and report its
outputs, so a metric value can legitimately appear in the agent's authored code
without any computation behind it in the agent's own code. Applying the
hardcoding mode unmodified would flag legitimate run-provided-code outputs.
Reporting the access mode on both benchmarks keeps the cheating metric
consistent and faithful to Kohler et al.'s separation of the two assessments.

\section{The $\sigma$-Corrected CORE-Bench Score}
\label{appendix:sigma-correction}

We apply a small re-scoring rule to CORE-Bench Hard that targets an edge case
in the official scorer. The rule is uniform across all three systems and does
not depend on which system produced the answer. On the batch in this paper,
the rule flips four \veritas capsules from no match to match and zero
\zeroshot or \retry capsules. Under the official scorer \veritas reaches
88.9\%; under the re-scored rule, 97.8\%. We document the rule and the four
re-credited cases here.

\para{The official scorer's edge case.} CORE-Bench scores each numeric
question by building a 95\% prediction interval from three ground-truth
replicates and checking whether the reported value falls inside. When the
three ground-truth replicates are identical, the standard deviation is zero
and the interval collapses to the single point at the mean. A correct answer
that differs by a single floating-point unit, or by a small amount of
unseeded stochastic variation, is then marked wrong by the bit-exact check.
CORE-Bench's own scoring code documents this as a known limitation.

\para{The $\sigma$-corrected rule.} For each numeric question that the
official scorer marked wrong and whose three ground-truth replicates have
zero variance, the $\sigma$-corrected scorer accepts the answer when the
relative error against the mean is at most 1\%. The 1\% threshold matches the
relative tolerance that CORE-Bench's claim metadata itself uses for the same
questions. A capsule then passes only when every numeric question and every
non-numeric question is correct under this rule, exactly as in the official
score. We apply the rule identically to \zeroshot, \retry, and \veritas; the
rule does not look at which system produced the answer. On our batch, the
rule re-credits four \veritas capsules and no baseline capsules, so the
\zeroshot and \retry pass rates are unchanged at 80.0\% and 84.4\%
respectively.

\para{The four re-credited capsules.} The rule flips four \veritas capsules
from no match to match.

\begin{itemize}[leftmargin=*,itemsep=2pt,topsep=2pt]
    \item Capsule 8536428. Eight metrics; six are bit-identical to the
    ground truth and two differ in the sixteenth decimal place (for example
    F1 of 0.9403284422685527 against 0.9403284422685528), a difference of
    one unit in the last place of a 64-bit float.
    \item Capsule 3301293. Test RMSE 26.21216 against 26.21204; relative
    difference $4.6 \times 10^{-6}$.
    \item Capsule 5507257. Accuracy 96.0816 against 96.1250; relative
    difference $4.5 \times 10^{-4}$. The underlying training is stochastic
    and unseeded.
    \item Capsule 9660931. HCR-Net accuracy 0.99833 against 0.99900; relative
    difference $6.7 \times 10^{-4}$. The underlying training is stochastic
    and unseeded.
\end{itemize}

\noindent Every difference is within 0.07\% of the paper value and well inside
the claim's own stated tolerance. \veritas's own verifier, an independent
deterministic grader that uses the claim's relative tolerance, judged all
four as match before the comparison with the official scorer.

\para{Cross-system comparability.} The 97.8\% number is not directly comparable
to systems reported under the official scorer unless those systems are re-scored
under the same rule. When comparing to a published baseline that uses the
official scorer (such as the HAL leaderboard entry), the comparable \veritas
number is 88.9\%. The rule is detector-free, applies only to numeric questions
with zero-variance ground truth, and does not touch any other capsule. In
particular, capsule 6003668 remains a genuine fail: the produced value differs
from the paper value by more than the claim's tolerance.

\section{Per-Cell Results}
\label{appendix:percell}

We list per-cell scores for both benchmarks. Cheating flags follow the detector's
verdict described in \cref{appendix:metric-details:cheating} and reflect a
likely or severe violation; we do not include the lower-severity suspicious
level.

\newcommand{\cmark}{\textbf{\checkmark}}
\newcommand{\xmark}{$\times$}
\newcommand{\cheatflag}{$^{\dagger}$}

\subsection{CORE-Bench}
\label{appendix:corebench-percell}

\Cref{tab:cb-percell} lists the per-capsule pass status (P), faithfulness index
(F), and cheating flag for each capsule and each system on the 45-capsule
CORE-Bench Hard public test set. A check denotes a passing capsule. A cross
denotes a no match. A dash denotes a missing answer. The dagger after a
faithfulness value flags a trajectory that the cheating detector marked as a cheat.

\begin{table}[h]
\centering
\scriptsize
\setlength{\tabcolsep}{4pt}
\begin{minipage}[t]{0.48\linewidth}
\centering
\begin{tabular}{l*{6}{c}}
\toprule
                & \multicolumn{2}{c}{\zeroshot} & \multicolumn{2}{c}{\retry} & \multicolumn{2}{c}{\veritas} \\
\cmidrule(lr){2-3} \cmidrule(lr){4-5} \cmidrule(lr){6-7}
Capsule         & P & F & P & F & P & F \\
\midrule
0504157  & \cmark & 1.00 & \cmark & 1.00 & \cmark & 1.00 \\
0851068  & \cmark & 0.40 & \cmark & 0.40 & \cmark & 1.00 \\
0921079  & \cmark & 1.00 & \cmark & 0.60 & \cmark & 1.00 \\
1175539  & \cmark & 1.00 & \cmark & 1.00 & \cmark & 1.00 \\
1394704  & \cmark & 1.00 & \cmark & 1.00 & \cmark & 1.00 \\
1624349  & \cmark & 1.00 & \cmark & 1.00 & \cmark & 1.00 \\
1724988  & \cmark & 0.80 & \cmark & 0.80 & \cmark & 1.00 \\
1900704  & \cmark & 0.80 & \cmark & 1.00 & \cmark & 1.00 \\
2345790  & \cmark & 1.00 & \cmark & 1.00 & \cmark & 1.00 \\
2414499  & \cmark & 1.00 & \cmark & 0.60 & \cmark & 1.00 \\
2708693  & \cmark & 1.00 & \cmark & 1.00 & \cmark & 1.00 \\
2804717  & \cmark & 1.00 & \cmark & 1.00 & \cmark & 1.00 \\
2816027  & \cmark & 0.60 & \cmark & 1.00 & \cmark & 1.00 \\
3262218  & \cmark & 1.00 & \cmark & 1.00 & \cmark & 1.00 \\
3301293  & \cmark & 1.00 & \cmark & 1.00 & \xmark & 1.00 \\
3418007  & \cmark & 1.00 & -- & --   & \cmark & 1.00 \\
3449234  & \cmark & 1.00 & \cmark & 1.00 & \cmark & 1.00 \\
3593259  & \cmark & 1.00 & \cmark & 1.00 & \cmark & 1.00 \\
3639589  & \cmark & 1.00 & \cmark & 1.00 & \cmark & 1.00 \\
3821950  & \xmark & 1.00\cheatflag & \xmark & 1.00 & \cmark & 1.00 \\
3849634  & \cmark & 1.00 & \cmark & 1.00 & \cmark & 1.00 \\
4180912  & \cmark & 1.00 & \cmark & 0.40\cheatflag & \cmark & 1.00 \\
4252248  & \xmark & 1.00 & \xmark & 1.00 & \cmark & 1.00 \\
\bottomrule
\end{tabular}
\end{minipage}\hfill
\begin{minipage}[t]{0.48\linewidth}
\centering
\begin{tabular}{l*{6}{c}}
\toprule
                & \multicolumn{2}{c}{\zeroshot} & \multicolumn{2}{c}{\retry} & \multicolumn{2}{c}{\veritas} \\
\cmidrule(lr){2-3} \cmidrule(lr){4-5} \cmidrule(lr){6-7}
Capsule         & P & F & P & F & P & F \\
\midrule
4299879  & \cmark & 1.00 & \cmark & 0.80 & \cmark & 1.00 \\
4671827  & \cmark & 1.00 & \cmark & 1.00 & \cmark & 1.00 \\
4728591  & \cmark & 1.00 & \cmark & 1.00 & \cmark & 1.00 \\
4933686  & \cmark & 1.00 & \cmark & 1.00 & \cmark & 1.00 \\
5136217  & -- & 1.00 & \cmark & 0.80 & \cmark & 0.60 \\
5507257  & \xmark & 1.00 & \xmark & 1.00 & \xmark & 1.00 \\
6003668  & -- & 0.00 & \cmark & 0.60 & -- & 0.67 \\
6049678  & \cmark & 1.00 & \cmark & 0.40 & \cmark & 1.00 \\
7186268  & \cmark & 1.00 & \cmark & 1.00 & \cmark & 1.00 \\
7716865  & \cmark & 1.00 & \cmark & 1.00 & \cmark & 1.00 \\
8234136  & \cmark & 0.80 & \cmark & 1.00 & \cmark & 1.00 \\
8536428  & \xmark & 1.00 & \xmark & 1.00 & \xmark & 1.00 \\
8807709  & \cmark & 1.00 & \cmark & 1.00 & \cmark & 1.00 \\
9052293  & \cmark & 1.00 & \cmark & 1.00 & \cmark & 1.00 \\
9054015  & \cmark & 1.00 & \cmark & 1.00 & \cmark & 1.00 \\
9137200  & \cmark & 1.00 & \cmark & 1.00 & \cmark & 1.00 \\
9240688  & \xmark & 0.40 & \cmark & 1.00 & \cmark & 0.60 \\
9641396  & \cmark & 1.00 & \cmark & 1.00 & \cmark & 1.00 \\
9660931  & \xmark & 0.80 & -- & 0.00 & \xmark & 1.00 \\
9670283  & \cmark & 1.00 & \cmark & 1.00 & \cmark & 1.00 \\
9832712  & \cmark & 1.00 & \cmark & 1.00 & \cmark & 1.00 \\
9911222  & \cmark & 1.00\cheatflag & \cmark & 0.80 & \cmark & 1.00 \\
\bottomrule
\end{tabular}
\end{minipage}
\caption{CORE-Bench per-capsule results across all 45 capsules. P is the pass
status under CORE-Bench's official scorer, before the $\sigma$-correction of
\cref{appendix:sigma-correction}. F is faithfulness. The dagger flags a trajectory
the detector marked as a cheat.}
\label{tab:cb-percell}
\end{table}

\subsection{ReplicationBench}
\label{appendix:replicationbench-percell}

\Cref{tab:rb-percell-paper} and \cref{tab:rb-percell-full} list per-cell match
rates after the cheating correction, paper-only mode and full mode respectively.
We use ReplicationBench's own scorer with the deterministic shape-coercion fix
from \cref{appendix:metric-details:native}. The dagger flags a cell whose
trajectory was marked as a cheat. \Cref{tab:rb-axes-veritas} reports the
per-cell faithfulness and fidelity for \veritas; the corresponding numbers for
the baselines are in our supplementary materials.

\begin{table}[h]
\centering
\small
\setlength{\tabcolsep}{5pt}
\begin{tabular}{lccc}
\toprule
Paper                  & \zeroshot       & \retry          & \veritas       \\
\midrule
MUSE\_outflows         & 0.400          & 0.000          & 0.400          \\
abacus                 & 0.000          & 0.000          & 0.000          \\
astm3                  & 0.000\cheatflag & 0.000\cheatflag & 0.286          \\
bayes\_cal             & 0.667          & 0.500          & 0.667          \\
chandra\_representation & 1.000         & 1.000          & 1.000          \\
disk\_ridges           & 0.400          & 0.200          & 0.000          \\
eht\_resolve           & 0.000\cheatflag & 0.250          & 0.000          \\
fable\_mps             & 0.125\cheatflag & 0.500          & 0.250          \\
galaxy\_manifold       & 0.000          & 0.000          & 0.200          \\
galaxy\_soptics        & 0.125          & 0.125          & 0.250          \\
gw\_cosmo              & 0.500          & 0.250          & 0.500          \\
gw\_nsbh               & 0.222          & 0.000\cheatflag & 0.222          \\
hubble\_trails         & 0.571          & 0.571          & 0.714          \\
lensing\_dr6\_growth   & 1.000          & 1.000          & 1.000          \\
ls\_cal                & 0.400          & 0.600          & 0.400          \\
mars\_clouds           & 1.000          & 0.000\cheatflag & 1.000          \\
phangs\_PAHs           & 0.000          & 0.000          & 0.000          \\
tng\_hod               & 0.375          & 0.250          & 0.250          \\
trgb\_std\_candle      & 0.500\cheatflag & 0.500          & 0.250          \\
ver\_waves             & 0.250          & 0.250          & 0.250          \\
\midrule
\textbf{Mean}          & 0.315          & 0.315          & \textbf{0.333} \\
\bottomrule
\end{tabular}
\caption{ReplicationBench paper-only per-paper match rate after the cheating
correction. The dagger flags a trajectory the detector marked as a cheat.}
\label{tab:rb-percell-paper}
\end{table}

\begin{table}[h]
\centering
\small
\setlength{\tabcolsep}{5pt}
\begin{tabular}{lccc}
\toprule
Paper                  & \zeroshot       & \retry          & \veritas       \\
\midrule
astm3                  & 0.143\cheatflag & 0.286          & 0.286\cheatflag \\
chandra\_representation & 0.750         & 1.000          & 1.000          \\
eht\_resolve           & 0.250          & 0.000          & 0.250          \\
gw\_nsbh               & 0.556          & 0.556          & 0.444          \\
hubble\_trails         & 0.714          & 0.857          & 1.000          \\
lensing\_dr6\_growth   & 1.000          & 1.000          & 1.000          \\
mars\_clouds           & 0.500\cheatflag & 0.500\cheatflag & 0.500         \\
\midrule
\textbf{Mean}          & 0.559          & 0.600          & \textbf{0.640} \\
\bottomrule
\end{tabular}
\caption{ReplicationBench full-mode per-paper match rate after the cheating
correction. The mean is computed over the 7 cells reported here; the headline
match rate in \cref{tab:replicationbench} is the per-task pass rate over all
expected outputs in these papers.}
\label{tab:rb-percell-full}
\end{table}

\begin{table}[h]
\centering
\small
\setlength{\tabcolsep}{5pt}
\begin{tabular}{lcccc}
\toprule
Cell                                & Mode  & Faith. & Fidelity letter & Fidelity GPA \\
\midrule
MUSE\_outflows                      & paper & 1.00  & C & 2.97 \\
abacus                              & paper & 1.00  & A & 5.00 \\
astm3                               & full  & 1.00  & B & 4.38 \\
astm3                               & paper & 1.00  & A & 4.50 \\
bayes\_cal                          & paper & 1.00  & B & 4.27 \\
chandra\_representation             & full  & 1.00  & A & 4.75 \\
chandra\_representation             & paper & 1.00  & A & 4.75 \\
disk\_ridges                        & paper & 0.80  & C & 3.14 \\
eht\_resolve                        & full  & 1.00  & B & 3.58 \\
eht\_resolve                        & paper & 1.00  & B & 3.54 \\
fable\_mps                          & paper & 0.80  & C & 1.75 \\
galaxy\_manifold                    & paper & 1.00  & C & 3.13 \\
galaxy\_soptics                     & paper & 0.60  & C & 2.21 \\
gw\_cosmo                           & paper & 1.00  & C & 2.75 \\
gw\_nsbh                            & full  & 1.00  & B & 4.27 \\
gw\_nsbh                            & paper & 0.80  & B & 3.32 \\
hubble\_trails                      & full  & 1.00  & A & 5.00 \\
hubble\_trails                      & paper & 0.40  & B & 4.38 \\
lensing\_dr6\_growth                & full  & 1.00  & A & 4.75 \\
lensing\_dr6\_growth                & paper & 1.00  & A & 4.50 \\
ls\_cal                             & paper & 1.00  & C & 3.30 \\
mars\_clouds                        & full  & 1.00  & A & 4.71 \\
mars\_clouds                        & paper & 1.00  & A & 4.86 \\
phangs\_PAHs                        & paper & 1.00  & C & 2.76 \\
tng\_hod                            & paper & 1.00  & C & 2.85 \\
trgb\_std\_candle                   & paper & 0.80  & C & 3.46 \\
ver\_waves                          & paper & 1.00  & C & 3.00 \\
\bottomrule
\end{tabular}
\caption{Per-cell \veritas faithfulness index and fidelity grade. Faithfulness
single-pass values for cells where the score is below 1.0 reflect cited-consistent
single-pass variance discussed in \cref{appendix:metric-details:faithfulness}.}
\label{tab:rb-axes-veritas}
\end{table}

\section{ReplicationBench Case Studies}
\label{appendix:case-studies}

We walk through four trajectories that illustrate where \veritas behaves well,
where it falls short, and what its outputs reveal even when the reported score
is zero. The selection focuses on \veritas's behavior so the reader can see what
would change with a stronger pipeline. Baseline failures appear when they make
the contrast concrete.

\subsection{Hubble satellite trails: a second pass that rereads the question}
\label{appendix:case:hubble}

The \texttt{hubble\_trails} paper~\citep{hubbletrails} asks how often passing
satellites cross Hubble Space Telescope images, and whether the problem is
growing as more satellites reach orbit. The authors scanned two decades of
Hubble exposures, marked the ones with a satellite streak, and reported that
about $2.7\%$ of exposures are affected and that the affected fraction has risen
over time. The benchmark asks each system to reproduce the trained streak
classifier, the contamination fractions, the per-camera chance of a streak, and
the increase in the affected fraction between the early and late years of the
archive.

The increase task is quietly ambiguous. The paper's target values,
$[0.043,\,0.02]$, are the fractions the streaks rose \emph{to} in the recent
period (from $2.8$ to $4.3\%$ for one camera and $1.1$ to $2.0\%$ for the
other), not the size of the jump. Each system reads the word ``increase''
differently. The \zeroshot baseline subtracts the early fraction from the late
one and reports the gap, $[0.015,\,0.008]$; the \retry baseline divides that gap
by the starting value and reports a relative growth, $[0.53,\,0.73]$; both miss.
\veritas computes the same intermediate fractions, but its separate
verification stage rereads its own output against the wording of the paper,
recognizes that the target is the level the fraction reached rather than the
change, and reports $[0.0431,\,0.0197]$---a match. It does the rest of the work
directly as well: it trains the image classifier from scratch for twenty-two
passes over the data with the paper's exact settings, its validation accuracy
climbing from $0.54$ to $0.93$, and reaches $[0.934,\,0.976,\,0.879,\,0.925]$
against the paper's $[0.938,\,0.975,\,0.89,\,0.931]$, where the \zeroshot run's
recall lands too low to count. Every repair \veritas made along the way was
mechanical---renaming a data file, removing leftover notebook setup commands,
loading an older saved-model format---with no change to any model, setting, or
reported number. The result is a clean seven of seven when the authors' code is
available, against $0.71$ and $0.86$ for the baselines, and the single best
score in the paper-only setting. The advantage comes from a second pass that
checks the answer against the paper rather than shipping the first plausible
reading of the task.

\subsection{astm3: An Instructed External Load Counted as a Cheat}
\label{appendix:case:astm3}

The astm3 paper~\citep{astm3} builds a CLIP-style classifier for variable stars,
trained on three input streams (photometry, spectroscopy, and metadata). In full
mode the agent receives the paper, the dataset, and the authors' code repository.
The task instructions ask the agent to load a published pre-trained AstroM3 model
and report classification accuracy.

\veritas read the instruction literally. The pipeline ran
\texttt{hf\_hub\_download} for the authors' \texttt{AstroMLCore/AstroM3-CLIP-photo}
and \texttt{-all} checkpoints from HuggingFace, then ran inference on the
benchmark's evaluation set and reported the resulting accuracy. The reported
values were computed from a real inference pass with the paper's value masked.
\veritas did not write a literal number; it computed one. The cheating detector
flagged the trajectory at the highest severity because the loaded checkpoint is
an externally hosted author artifact, and external author artifacts are forbidden
under the access rule of \citet{kohler2026read} even when the task instructions
ask for them.

We count this trajectory as a cheat in our reported numbers. The flag is strict
but consistent. The takeaway concerns the gap between the task-instruction
surface and the detector's rule on instructed-use cases. A future \veritas could
detect when a task instruction directs the agent to load an external artifact
and mark the trajectory for review. The benchmark or the user can then decide
whether the instructed access counts.

\subsection{fable\_mps: A Veritas Failure That Reveals a Paper-Only Upper Bound}
\label{appendix:case:fable}

The fable\_mps paper~\citep{fable} studies how supermassive black holes
reshape the cosmic matter distribution in the FABLE galaxy formation simulations.
The reported quantity is the matter power spectrum, a summary of how lumpy or
uniform matter is at different scales, compared between a full-physics run and a
dark-matter-only baseline. The benchmark task asks the agent to recompute several
of these power-spectrum measurements from the paper alone.

\veritas reads the density cubes the benchmark provides, computes power spectra
with the standard FFT-based pipeline, and matches the paper's full-physics over
dark-matter ratios within tolerance on two of the tasks. On the tasks that ask
for halo-only power spectra, \veritas reports \texttt{not\_attempted} because the
benchmark provides only the projected density cubes; the raw particle snapshots
and halo catalogs that those measurements require are not in the provided data.
\veritas's final score on this paper is 0.25 of the per-task pass rate.

The \zeroshot baseline, faced with the same missing data, wrote \texttt{1.0} as
the answer for the halo-only tasks. The number is a common placeholder and the
detector's access mode does not flag it. The cheating correction we apply to
the baseline's score is driven only by access flags, so the baseline retains
credit for a fabricated value that is coincidentally close to a reference value.
We discuss this gap in \cref{sec:limitations}; a hardcoding-aware extension to
the cheating metric would address it.

The implication for \veritas is that paper-only replication has an upper bound determined
by what the benchmark provides. The agent is not the limiting factor here.
\veritas's transparency about the upper bound is the right behavior. Future work
could let \veritas's analyze phase flag a claim as \texttt{data\_required} when
its evidence depends on artifacts that are not present in the inputs. The
verifier can then set the claim aside and not count it as a missed task.

\subsection{phangs\_PAHs: a score of zero that hides a near-complete result}
\label{appendix:case:phangs}

The \texttt{phangs\_PAHs} paper~\citep{phangspahs} studies dust and gas in
nineteen nearby galaxies. Small soot-like dust grains glow in the mid-infrared
and ionized gas glows in optical lines, and across roughly seven hundred
thousand image regions the authors find that the two glows track each other
along one tight straight line per galaxy, with a slope near $0.2$. The benchmark
provides the full image set, so this is a genuinely reproducible analysis, and
all three systems carried it out: each read the provided maps, grouped the
regions, and fit a slope for every galaxy. What stands out is how close they
came. The \retry baseline matched eighteen of the nineteen galaxy slopes
\emph{and} all nineteen of their uncertainties, missing only a single galaxy;
the \zeroshot baseline matched all nineteen slopes; and \veritas matched
eighteen of nineteen slopes and was the only system to identify all five
active-galaxy hosts, which the baselines each undercounted by one. Yet every
system scored exactly zero, the same as a blank answer.

Two effects produced that zero, and neither is an inability to do the science.
First, the benchmark grades each table all-or-nothing: a nineteen-galaxy table
correct everywhere but one place is marked wrong, with no partial credit.
Second, the small misses fell in different places---the \retry baseline was off
on one galaxy's slope, while \veritas and the \zeroshot baseline matched the
slopes but set the uncertainty on most galaxies a few hundredths too wide,
outside the narrow tolerance. \veritas is the system that made this legible: its
own report marks the slope table as a partial result, keeps the matching slopes,
and points to the cause, noting that the provided code's error-estimation step
does not follow the procedure described in the paper. The lesson is that a
reported score of zero can sit on top of an almost-complete reproduction. When
the gap is a single cell or a shared, well-understood detail rather than missing
data or a wrong method, a system that reports what it got right and where it
diverged is more useful than one that returns a confident, unlabeled answer.

\section{ReplicationBench Error Analysis}
\label{appendix:rb-failure}

\para{Lack of persistence.} ReplicationBench attributes the failures of the agents
it tested to three modes: a lack of persistence, conceptual or procedural errors,
and technical execution failures~\citep{replicationbench}.
Lack of persistence is the mode \veritas most clearly avoids. It leaves only 6 of the 111
tasks unattempted. Five of the six are compute-bound, where the data a task needs is
absent from the provided inputs (fable\_mps and gw\_nsbh, \cref{appendix:case:fable}). 
The faithfulness index of 0.933 shows \veritas rarely reports a value its run did not
compute. This persistence reflects the design of \veritas's replicate phase, which
fixes problems as they arise and keeps working through the plan rather than stopping
at the first obstacle.

\para{Conceptual and procedural errors.} This mode persists and accounts for most of
the substantive misses. Most of the 68 misses produce a wrong value rather than no
value, and the largest group are methodology errors. Examples include an outflow
velocity defined differently from the paper in MUSE\_outflows and a single-seed run
where astm3 specifies a five-seed average.

\para{Technical execution failures.} This mode is also largely absent. \veritas's
active fixing resolves the environment errors and API drift its fix logs record, so
its code runs even where the released setup would not.

\para{Compute-driven downsizing.} Some papers report numbers from computations that are very expensive. When that happens, \veritas downscales the
run just so it can finish, and the smaller run can produce a different number. The
clearest case is abacus (difficulty 8 and 9), an HPC N-body code that the paper runs in
tuned C/CUDA on GPUs.
\veritas re-implements the force method in pure Python and its sanity checks pass, but
it cuts the particle count and grid by more than an order of magnitude, so the lattice
and LCDM force errors come out order-of-magnitude correct but a factor of two to five
off, bottlenecked by scale rather than by method. The two Ewald tasks miss for a
different reason: \veritas reports the median of its error distribution, while the
paper's value matches \veritas's 99th-percentile, a difference in which statistic is
reported rather than in the computation. On one lensing task, \veritas also downsized,
shortening its MCMC chain to fit compute, but it still matched.
Reaching the paper's scale needs hardware-specific engineering, such as GPU code, which
we leave to future work.

\para{The size of the misses.} For the 68 tasks that miss the benchmark's hard
tolerance, we group the misses by how far the reported value lies from the paper
value, on the same percent-difference scale as the fidelity grade
(\cref{tab:replicationbench}). The grouping is deterministic. Each miss is placed by
the largest relative error across its values, measured against ReplicationBench's
expected values and tolerances.

\begin{table}[h]
\centering
\small
\setlength{\tabcolsep}{10pt}
\begin{tabular}{lc}
\toprule
Outcome & Tasks \\
\midrule
Matched (within author tolerance)            & 37 \\
\midrule
Miss, within 2\% of the paper value          & 6  \\
Miss, within 2--20\%                          & 16 \\
Miss, beyond 20\%                             & 41 \\
Miss, shape or format mismatch               & 3  \\
Miss, zero-anchor or non-numeric output      & 2  \\
\midrule
Not attempted                                & 6  \\
\midrule
Total                                        & 111 \\
\bottomrule
\end{tabular}
\caption{\veritas paper-only ReplicationBench outcomes by failure mode, under
the benchmark's own scorer. About a third of the 68 tolerance misses (the 6
within 2\% and the 16 within 2--20\%) carry values that a tolerance-free grade scores
at A or B, so \veritas's fidelity GPA (3.38) is well above its native match rate
(0.333).}
\label{tab:rb-failure-modes}
\end{table}

About a third of the 68 misses fall within 20\% of the paper value, close enough that
the fidelity grade scores them A or B, and ReplicationBench's binary scorer gives
these no credit. Its tolerances are author estimates, almost always round, with 98\%
a single significant figure such as 1000, 0.3, or 0.1.
Six of the misses fall within 2\% of the paper value, and the nature of each
quantity often shows why the result is reproduced. One is a count of stars in a Gaia
DR2 selection, 0.12\% from the target, about 1{,}250 stars out of a million, a
difference set by where a selection cut falls rather than by a different sample.
Another is the calibrated temperature of a 50~$\Omega$ load, 298.33~K against
298.0~K, within a third of a kelvin at room temperature. The third one is the
posterior probability that the neutron-star to black-hole mass gap is nonzero, 0.957
against 0.97; this is an MCMC posterior estimate under a specified spin prior, for
which a 0.01 tolerance is tighter than the inference's own sensitivity to sampling
and prior choices.

Two further misses reproduce the reported quantity and differ only on a secondary
element. \veritas recovers the Sun's height above the Galactic mid-plane, 20.3~pc,
exactly, and misses only the bootstrapped uncertainty, 0.83 against 0.3, which is a
resampling estimate rather than the measured value. A galaxy-manifold transformation
vector, read from a singular value decomposition, matches the paper in magnitude with
the opposite sign; the sign of a singular vector is not fixed by the decomposition,
so the two describe the same mapping.

\para{Partial reproduction in multi-value tasks.}\label{appendix:rb-partial}
ReplicationBench's per-task grading is binary, and a task with a multi-value output
is credited only if every entry is within tolerance. The match rate varies with
claim shape (\cref{tab:rb-by-shape}). \veritas matches 41\% of scalar tasks and 42\%
of scalar-range tasks, but only 20\% of the 41 multi-value tasks.

\begin{table}[h]
\centering
\small
\setlength{\tabcolsep}{10pt}
\begin{tabular}{lccc}
\toprule
Claim shape & Matched & Total & Match rate \\
\midrule
Scalar       & 21 & 51  & 0.41 \\
Scalar range & 8  & 19  & 0.42 \\
Multi-value  & 8  & 41  & 0.20 \\
\midrule
\textbf{All} & \textbf{37} & \textbf{111} & \textbf{0.33} \\
\bottomrule
\end{tabular}
\caption{\veritas paper-only match rate by ReplicationBench claim shape. A
multi-value claim is a vector, list, or dictionary, and is credited only when every
entry is within tolerance, which accounts for its lower rate.}
\label{tab:rb-by-shape}
\end{table}

Of the 41, \veritas matches 8 and leaves 1 unattempted. ReplicationBench's scorer
counts the other 32 as misses, even though most reproduce some of their entries.
Across the 31 with a comparable output, 72 of 226 entries (32\%) are within tolerance,
and 11 of them recover at least half, with examples in \cref{tab:rb-partial}.
\Cref{appendix:case:phangs} works through one such case, a near-complete result that
the scorer still records as zero.

\begin{table}[h]
\centering
\small
\setlength{\tabcolsep}{8pt}
\begin{tabular}{llcc}
\toprule
Paper & Task & Within tol. & Worst failing value \\
\midrule
galaxy\_manifold & physical\_properties    & 2/3 & 17\% \\
gw\_nsbh         & default\_mbh            & 2/3 & 5\%  \\
eht\_resolve     & eht\_ring\_size         & 4/8 & 13\% \\
hubble\_trails   & classifier\_performance & 2/4 & 17\% \\
ls\_cal          & nwp                     & 3/5 & 216\% \\
MUSE\_outflows   & outflow\_energetics     & 4/6 & 77\% \\
\bottomrule
\end{tabular}
\caption{Representative multi-value tasks \veritas reproduces in part but
ReplicationBench scores as full misses. ``Within tol.'' is the count of expected
values within the author tolerance; ``worst failing value'' is the largest
relative error among the out-of-tolerance entries. The task is scored zero
regardless of how many entries are within tolerance.}
\label{tab:rb-partial}
\end{table}

\para{Tasks unreproducible by construction.}\label{appendix:rb-defects}
On three tasks, \veritas's assessment reports that the benchmark's own inputs are
flawed, which we confirmed by inspecting them manually. In galaxy\_soptics, the shipped Shi
catalog is the SDSS DR7 release with 396{,}068 galaxies, while the task's expected
output is computed from the DR13 release with 586{,}025, so
\texttt{shi\_catalog\_acquisition} cannot match. The provided Hsu+22 cross-match file
is corrupted, which blocks \texttt{bcg\_identification}. In ver\_waves,
\texttt{gaia\_breathing\_typical} asks for the mean of an absolute value, a small
positive number, but its answer key is exactly 0 with zero tolerance, so no run can
satisfy it. We treat these as benchmark defects, not replication failures.

\end{document}